\documentclass{article}
\usepackage{ijcai25}

\usepackage{times}
\usepackage{soul}
\usepackage{url}
\usepackage[hidelinks]{hyperref}
\usepackage[utf8]{inputenc}
\usepackage[small]{caption}
\usepackage{graphicx}
\usepackage{amsmath}
\usepackage{amsthm}
\usepackage{booktabs}
\usepackage{algorithm}
\usepackage{algorithmic}
\usepackage[switch]{lineno}
\usepackage{color}
\usepackage{xcolor}
\usepackage{bbm}
\usepackage{amsfonts}
\usepackage{cleveref}
\usepackage{multirow}
\usepackage{multicol}
\usepackage{array}

\newcommand{\tang}[1]{{\color{red}{#1}}}

\newcommand{\wang}[1]{{\color{red}{Xiaofei: #1}}}
\newcommand{\OurModel}[1]{AdvGrasp}
\newcommand{\firstpara}[1]{\noindent\textbf{{#1}.}~~}

\title{\OurModel{}: Adversarial Attacks on Robotic Grasping from a Physical Perspective
}

\author{
Xiaofei Wang$^{1,2}$
\and
Mingliang Han$^1$\and
Tianyu Hao$^{3}$\and
Cegang Li$^1$\and
Yunbo Zhao$^{1,4\ast}$\And
Keke Tang$^{3}$\thanks{Yunbo Zhao and Keke Tang are co-corresponding authors.}
\affiliations
$^1$Department of Automation, University of Science and Technology of China\\
$^2$SmartMore Corporation\\
$^3$Cyberspace Institute of Advanced Technology, Guangzhou University\\
$^4$Institute of Artificial Intelligence, Hefei Comprehensive National Science Center\\
\emails
wxf9545@mail.ustc.edu.cn, mlhan@mail.ustc.edu.cn, howty666@gmail.com, \\lcg123@mail.ustc.edu.cn, 
ybzhao@ustc.edu.cn, tangbohutbh@gmail.com
}

\begin{document}

\maketitle

\begin{abstract}

\if 0
Adversarial attacks on robotic grasping provide valuable insights for evaluating and enhancing the robustness of these systems.
Unlike studies focused on neural network predictions while neglecting the physical principles of grasping, this paper introduces \OurModel{}, a framework for adversarial attacks on robotic grasping from a physical perspective.
Specifically, \OurModel{} addresses two core aspects: lift capability, which assesses the ability to lift objects against gravity, and grasp stability, which measures resistance to external disturbances.
By deforming the object's shape to  increase the gravitational torque and 
reduce the stability margin in the wrench space,
our method systematically degrades the two key grasping metrics, creating adversarial objects that challenge grasp performance.
Extensive experiments across diverse scenarios validate the effectiveness of \OurModel{}, while real-world validations confirm its robustness and practical applicability. Codes and benchmarks will be released upon paper acceptance.
\fi

Adversarial attacks on robotic grasping provide valuable insights into evaluating and improving the robustness of these systems.  
Unlike studies that focus solely on neural network predictions while overlooking the physical principles of grasping, this paper introduces \OurModel{}, a framework for adversarial attacks on robotic grasping from a physical perspective.  
Specifically, \OurModel{} targets two core aspects: lift capability, which evaluates the ability to lift objects against gravity, and grasp stability, which assesses resistance to external disturbances.  
By deforming the object's shape to increase gravitational torque and reduce stability margin in the wrench space, our method systematically degrades these two key grasping metrics, generating adversarial objects that compromise grasp performance.  Extensive experiments across diverse scenarios validate the effectiveness of \OurModel{}, while real-world validations demonstrate its robustness and practical applicability. 

%Codes and benchmarks will be released upon paper acceptance. % https://github.com/ustcwxf/advgrasp

%perturb its center of mass and surface normals, 

\end{abstract}

\section{Introduction}

Grasping serves as a fundamental mechanism for robots to interact with the physical world~\cite{844081,lin2022manipulation}, enabling a wide range of applications from industrial automation~\cite{cutkosky1989grasp} to domestic services~\cite{matheus2010benchmarking}. As a critical component of many safety-critical systems, even minor errors in robotic grasping can lead to severe and unpredictable consequences. Recent studies have shown that robotic grasping systems are vulnerable to various threats~\cite{yaacoub2022robotics}, including adversarial attacks~\cite{szegedy-2014-intriguing}, where imperceptible perturbations to input data can cause erroneous decisions. Understanding and addressing these threats, including adversarial attacks, is essential for improving the robustness and reliability of robotic grasping systems.

\begin{figure}[!t]
    \centering
    \includegraphics[width=0.88\linewidth]{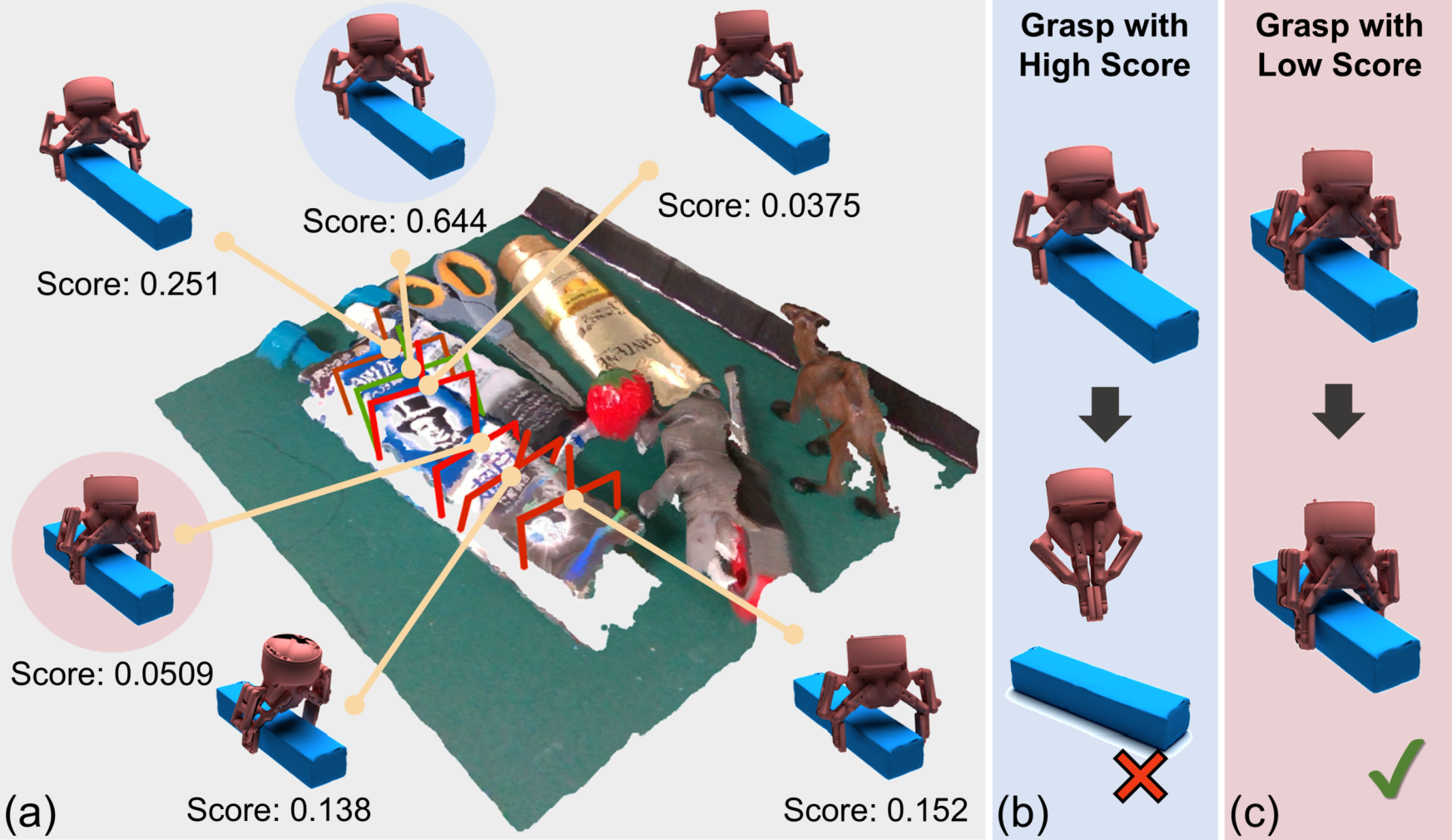}
    \caption{
Illustration of the limitations of relying solely on neural network predictions.
(a) A scenario from GraspNet-1Billion, where the provided baseline network, i.e., GraspNet, generates multiple grasps with quality scores~\protect\cite{fang2020graspnet,fang2023graspnet2}.
 However, (b) a predicted grasp with a high-quality score fails in execution, while (c) another grasp with a low-quality score succeeds.
% Visualization of a scenario from GraspNet-1Billion, where the GraspNet-Baseline algorithm generates multiple grasps. (a) depicts the initial scene and generated grasps; (b) shows a high-quality predicted grasp that fails in practice, while (c) illustrates a low-quality predicted grasp that succeeds. These results underscore the limitations of relying solely on neural network predictions without considering the physical dynamics of robotic grasping.
%    Given (a) a GraspNet-1Billion的场景上, GraspNet-Baseline 算法生成了一些grasp；    we observe that (b) a grasp with a high predicted quality fails, while (c) a grasp with a low predicted quality succeeds, highlighting the limitations of considering the physical process of grasping purely from a neural network perspective.
    }
    \label{fig:teaser}
\end{figure}

%我们使用GraspNet-Baseline 算法在GraspNet-1Billion的场景上

%Despite these developments, research on adversarial attacks in robotic grasping remains relatively limited. \citet{alharthi2024physical} extended adversarial attack techniques from image-based classification neural networks to image-based grasp quality networks, generating adversarial examples that reduced predicted grasp quality. However, these attacks focus solely on exploiting vulnerabilities in grasp quality networks without addressing the physical grasping process directly. Fundamentally, this approach targets a neural network rather than the physical act of grasping. One limitation is that not all robotic grasping systems rely on neural networks for grasp quality evaluation, which limits the applicability of such attacks. Moreover, even in systems that do use these networks, their instability under adversarial conditions highlights the unreliability of their predictions. Successfully attacking a grasp quality network does not necessarily translate to disrupting the physical grasp itself, see Fig.~\ref{fig:teaser}.  This highlights the importance of developing adversarial attack strategies that directly target the grasping process, accounting for the physical interactions and challenges inherent in robotic grasping.

Despite these developments, research on adversarial attacks in robotic grasping remains relatively limited. Alharthi and Brand{\~a}o~\shortcite{alharthi2024physical} extended adversarial attack techniques from image-based classification neural networks to image-based grasp quality networks, generating adversarial examples that reduced predicted grasp quality. However, these attacks focus solely on exploiting vulnerabilities in grasp quality networks without addressing the physical grasping process directly. Fundamentally, this approach targets a neural network rather than the physical act of grasping. One limitation is that not all robotic grasping systems rely on neural networks for grasp quality evaluation, which limits the applicability of such attacks. Moreover, even in systems that do use these networks, their instability under adversarial conditions highlights the unreliability of their predictions. Namely, successfully attacking a grasp quality network does not necessarily translate to disrupting the physical grasp itself, see Fig.~\ref{fig:teaser}.
This highlights the importance of developing adversarial attack strategies that directly target the grasping process, accounting for the physical interactions and challenges inherent in robotic grasping.

Indeed, robotic grasping is a fundamentally complex physical process rather than a straightforward end-to-end task. It requires intricate interactions involving forces, torques, and friction, which are difficult to model solely with neural networks. Furthermore, a successful grasp involves multiple stages: overcoming gravity to lift the object and maintaining stability throughout transportation, often under the influence of external disturbances such as vibrations, collisions, or unexpected forces. These challenges underscore the limitations of adversarial attacks that fail to consider the physical realities of grasping and instead focus exclusively on neural network vulnerabilities. This raises a critical question: can adversarial attacks on robotic grasping be systematically formulated from a physical perspective, capturing these essential interactions and processes?

To address these challenges, we devise \OurModel{}, a systematic adversarial attack framework for robotic grasping from a physical perspective.
Specifically, \OurModel{} focuses on two critical aspects of grasp performance: lift capability, which evaluates the gripper's ability to overcome gravity, and grasp stability, which measures its capacity to resist external disturbances.   
By subtly deforming the object's shape to 
 increase the gravitational torque and 
reduce the stability margin in the wrench space,
%modify its center of mass and surface normals,
our method disrupts the grasp wrench equilibrium, reducing the physical effectiveness of grasping.
To validate our approach, we introduce AdvGrasp-20, a comprehensive benchmark featuring 20 groups of objects with diverse shapes and multiple grasp configurations generated by both traditional and neural network-based methods for common two- and three-finger robotic grippers.
Extensive experiments across various settings on this benchmark demonstrate the effectiveness of \OurModel{}. Besides, we also validate its performance in real-world scenarios to further confirm its  practical applicability in physical robotic systems.

%Extensive experiments across various settings on this benchmark demonstrate the effectiveness of \OurModel{}. Additionally, real-world validations confirm its robustness and practical applicability in physical robotic systems.

%To validate our approach, we introduce AdvGrasp-20, a comprehensive benchmark featuring 20 object groups with diverse shapes and multiple grasp configurations generated by both traditional and neural network-based methods for common two- and three-finger robotic grippers. 

%To address these challenges, we propose \OurModel{}, a framework that systematically formulates adversarial attacks on robotic grasping from a physical perspective. Specifically, \OurModel{} targets two key aspects of grasp performance: lift capability, which evaluates the gripper's ability to overcome gravity, and grasp stability, which measures its capacity to resist external disturbances. By subtly altering the object's center of mass and surface normals, our method disrupts the grasp wrench equilibrium and degrades the physical effectiveness of grasping. To validate our approach, we construct a comprehensive adversarial grasping benchmark, AdvGrasp-20, comprising 20 object groups with diverse shapes and multiple grasp configurations generated by both traditional and neural network-based methods for two- and three-finger grippers. Extensive experiments on this benchmark demonstrate the effectiveness of \OurModel{} under various settings. Furthermore, validations in real-world scenarios confirm its robustness and practical applicability.

Overall, our contribution is summarized as follows:
\begin{itemize}

    \item We present AdvGrasp-20, a benchmark with diverse object shapes and grasp configurations to standardize the evaluation of adversarial attacks on grasping.
    
    \item We propose a physical-aware adversarial attack framework for robotic grasping that systematically targets two critical metrics: lift capability and grasp stability.
    
    \item We validate the proposed framework through extensive experiments and real-world evaluations, demonstrating its effectiveness in degrading grasp performance.
    
\end{itemize}

\section{Related Work}

\subsection{Security Issues in Robotics}

As robotic systems are applied more broadly in diverse applications~\cite{chen2015intelligent,siciliano2008springer}, their interconnected modules, such as communication networks, perception systems, and control mechanisms, have increasingly become targets for various attacks~\cite{yaacoub2022robotics}. 
%These vulnerabilities can disrupt system functionality, compromise safety, and undermine overall reliability.
Network security threats, such as data injection and denial-of-service (DoS), have been shown to compromise localization and disrupt operations~\cite{guerrero2018detection}. Similarly, tampering with sensor data or control signals in human-robot collaboration can lead to unsafe behaviors or task failures~\cite{meng2021dynamic,amaya2022vision}.
Operating systems like the Robot Operating System (ROS) are also vulnerable. Dieber \emph{et al.}~\shortcite{dieber2017security} identified key weaknesses in ROS, while Mazzeo and Staffa~\shortcite{mazzeo2020tros} proposed methods to mitigate privileged-access attacks.
Although robotic security has been widely studied, adversarial attacks specifically targeting robotic grasping remain largely unexplored. This paper aims to bridge this gap by systematically investigating such attacks.

\subsection{Adversarial Attacks on Robotic Grasping}

Adversarial attacks aim to generate perturbed examples that mislead neural networks into making incorrect predictions. Since the introduction of adversarial examples by Szegedy \emph{et al.}~\shortcite{szegedy-2014-intriguing}, these attacks have been extensively studied in computer vision~\cite{Goodfellow-2014-FGSM,Dong-2018-MIFGSM,Carlini-2017-cw,Moosavi-2016-deepfool,tang2024effective} and extended to domains such as 3D point cloud perception~\cite{Xiang-2019-3DAdversarialPCD,tang2022rethinking,Tang-2023-ManifoldAttack,tang2024manifoldConstraints,Tang-2024-FLAT,tang2024symattack,tang2025imperceptible,tang2025imperceptibleICASSP,wang2025eia}, natural language processing~\cite{morris2020textattack}, and audio processing~\cite{zheng2021black}. Recently, robotic grasping has also been investigated within the context of adversarial attacks.

Alharthi and Brand{\~a}o~\shortcite{alharthi2024physical} extended adversarial attack techniques from image-based classification neural networks to image-based grasp quality networks. Their work demonstrated adversarial examples in robotic grasping by modifying the intensity of a single pixel or physically placing a barely visible spherical object. However, this approach focuses on attacking the grasp quality network itself, rather than addressing the grasping process as a whole.
Grasping is inherently a complex physical interaction involving factors such as forces, torques, and friction. Therefore, effective adversarial attacks on robotic grasping must account for these physical characteristics to accurately capture the challenges of the task.

Wang \emph{et al.}~\shortcite{wang2019adversarial} incorporated certain physical properties of grasping in their study, but their objective was to generate objects that are universally difficult to grasp from any angle. This objective significantly increases the complexity of optimization and often leads to adversarial objects with unrealistic and conspicuous deformations. In contrast, our work focuses on attacking specific grasps of an object by systematically investigating adversarial attacks from a physical perspective, while ensuring that the generated perturbations remain realistic and physically plausible.

\begin{figure*}[!t]
    \centering
\includegraphics[width=0.92\linewidth]{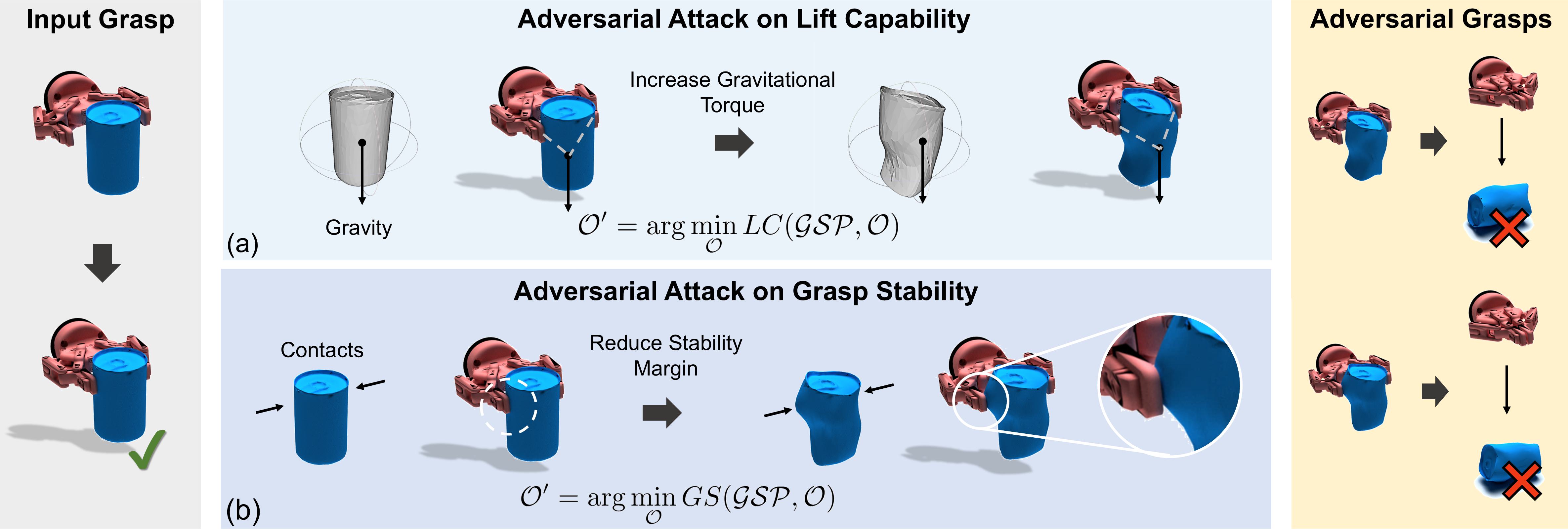}
    \caption{
%Illustration of \OurModel{}: Given a 3D object and its corresponding grasp configuration as inputs, \OurModel{} deforms the object in two key aspects: (a) increase gravitational torque to reducing lift capacity and (b) reduce stability margin in the wrench space to compromise grasp stability, ultimately causing grasp failure.
Illustration of \OurModel{}:
Given a 3D object and its corresponding grasp configuration as inputs, \OurModel{} deforms the object in two key ways: (a) increasing gravitational torque to weaken lift capacity and (b) reducing the stability margin in the wrench space to compromise grasp stability, ultimately leading to grasp failure.
}
    \label{fig:framework}
\end{figure*}

\section{Problem Statement}

\subsection{Preliminaries}

Formally, a 3D object $\mathcal{O}$ is represented as a triangle mesh defined by a set of triangular faces $\mathcal{T}(\mathcal{O})$ and vertices $\mathcal{V}(\mathcal{O})$. A successful grasp on $\mathcal{O}$ is denoted as $\mathcal{GSP} = \{\{\mathbf{c}_i\}, \{\mathbf{f}_i\}\}$, where $\{\mathbf{c}_i\}$ are the contact points, and $\{\mathbf{f}_i\}$ are the forces exerted by the robotic fingers.

At each contact point $\mathbf{c}_i$, the force $\mathbf{f}_i$ can be decomposed into two components:
\[
\mathbf{f}_i = \mathbf{f}_i^\perp + \mathbf{f}_i^t,
\]
where \( \mathbf{f}_i^\perp \) is the normal component, and \( \mathbf{f}_i^t \) is the tangential component. According to the Coulomb's law~\cite{mason1989robot}, the tangential force is constrained by the normal force:
\[
\|\mathbf{f}_i^t\| \leq \mu \|\mathbf{f}_i^\perp\|,
\]
where $\mu$ is the coefficient of friction.

\subsection{Stable Condition of A Grasp}
\label{subsec:Stable}

\firstpara{Wrench}
A wrench $\mathbf{w} \in \mathbb{R}^{1\times6}$ is a vector that encapsulates both force and torque, defined as:
\[
\mathbf{w} =
\begin{pmatrix}
\mathbf{f}, \
\mathbf{\tau}
\end{pmatrix},
\]
where $\mathbf{f} \in \mathbb{R}^{1\times3}$ represents the force vector, and $\mathbf{\tau} \in \mathbb{R}^{1\times3}$ denotes the 
torque vector acting on the object.

\if 0
\firstpara{Grasp Wrench Equilibrium}
For a grasp $\mathcal{GSP}$ on an object $\mathcal{O}$ to be stable, the wrenches applied by the robotic gripper must exactly counteract the external wrench $\mathbf{w}_{external}$, which may arise from a single force or a combination of multiple forces. This equilibrium condition is expressed as:
\begin{equation}
\sum \mathbf{w}_i + \mathbf{w}_{external} = \mathbf{0},
\label{eq:wrench_external}
\end{equation}
where $\mathbf{w}_i$ is the wrench applied by the gripper at a contact point $\mathbf{c}_i$, defined as:
\begin{equation}
\mathbf{w}_i = 
\begin{pmatrix}
\mathbf{f}_i \\
(\mathbf{c}_i - \mathbf{z}) \times \mathbf{f}_i + \tau_i
\end{pmatrix},
\label{eq:wrench_balance}
\end{equation}
with
\[
\tau_i \leq \mu \cdot \gamma \cdot \mathbf{f}_i^\perp\|
\]
where $\tau_i$ 是soft finger contact model~\cite{mahler2018dex}里的rotational torque, 
$\gamma$ is torsion friction coefficient
and  $\mathbf{z}$ denoting the centroid of the object.
\fi

\firstpara{Grasp Wrench Equilibrium}
For a grasp $\mathcal{GSP}$ on an object $\mathcal{O}$ to be stable, the wrenches applied by the robotic gripper must counteract the external wrench $\mathbf{w}_{\text{external}}$, which may result from a single or multiple forces. This grasp wrench equilibrium is expressed as:
\begin{equation}
\sum \mathbf{w}_i + \mathbf{w}_{\text{external}} = \mathbf{0},
\label{eq:wrench_external}
\end{equation}
where $\mathbf{w}_i$ is the wrench exerted by the gripper at contact point $\mathbf{c}_i$.
In a coordinate system centered at the object’s centroid $\mathbf{z}$, $\mathbf{w}_i$ is defined as:
\begin{equation}
\mathbf{w}_i =
\begin{pmatrix}
\mathbf{f}_i, \
(\mathbf{c}_i - \mathbf{z}) \times \mathbf{f}_i + \tau_i
\end{pmatrix},
\label{eq:wrench_balance}
\end{equation}

For soft finger contact models~\cite{mahler2018dex}, the rotational torque $\tau_i$ satisfies:
\[
\|\tau_i\| \leq \gamma  \|\mathbf{f}_i^\perp\|,
\]
where $\gamma$ is the torsional friction coefficient.

% \firstpara{Grasp Wrench Equilibrium}
% For a grasp $\mathcal{GSP}$ on an object $\mathcal{O}$ to be stable, the wrenches applied by the robotic gripper must fully counteract the external wrench $\mathbf{w}_{\text{external}}$, which may originate from a single force or a combination of multiple forces. This grasp wrench equilibrium condition is formally expressed as:
% \begin{equation}
% \sum \mathbf{w}_i + \mathbf{w}_{\text{external}} = \mathbf{0},
% \label{eq:wrench_external}
% \end{equation}
% where $\mathbf{w}_i$ represents the wrench exerted by the gripper at the contact point $\mathbf{c}_i$. Specifically, 以物体质心 $(\mathbf{z})$构建坐标系，$\mathbf{w}_i$ is defined as:
% \begin{equation}
% \mathbf{w}_i = 
% \begin{pmatrix}
% \mathbf{f}_i \\
% (\mathbf{c}_i - \mathbf{z}) \times \mathbf{f}_i + \tau_i
% \end{pmatrix},
% \label{eq:wrench_balance}
% \end{equation}
% where $\mathbf{f}_i$ is the contact force at $\mathbf{c}_i$. The term $\tau_i$ represents the rotational torque at the contact point. For the soft finger contact model~\cite{mahler2018dex}, the rotational torque $\tau_i$ satisfies the following constraint:
% \begin{equation}
% \|\tau_i \|\leq  \gamma \mu \|\mathbf{f}_i^\perp\|,
% \end{equation}
% where $\gamma$ is the  torsional friction coefficient.

% This formulation accounts for both the forces and torques applied at each contact point, ensuring that their combined effect counterbalances the external wrench $\mathbf{w}_{\text{external}}$ to maintain the stability of the grasp.

This condition ensures that the external wrench is fully balanced by the combined wrenches applied at the contact points by the robotic gripper, thereby maintaining grasp stability.

\subsection{Adversarial Attacks on Robotic Grasping}

%The goal of adversarial attacks on robotic grasping is to degrade the performance of a given grasp configuration \( \mathcal{GSP} \) by perturbing the object \( \mathcal{O} \), producing an adversarial version \( \mathcal{O}' \). 

The goal of adversarial attacks on robotic grasping is to degrade the performance of a given grasp configuration \( \mathcal{GSP} \) on an object \( \mathcal{O} \) by perturbing it, producing an adversarial version \( \mathcal{O}' \).
These attacks can be approached from two core aspects of grasp functionality:
\begin{itemize}
    \item \textbf{Lift Capability}: The gripper's ability to overcome gravity and successfully lift the object. 
    \item \textbf{Grasp Stability}: The gripper's ability to maintain a secure hold under external disturbances. 
\end{itemize}

By degrading these two core aspects, adversarial attacks effectively compromise robotic grasping performance.

\section{Method}

%In this section, we present \OurModel{}, a framework for systematically generating adversarial objects to degrade robotic grasping performance by targeting lift capability and grasp stability. We detail the key components and attack strategies, followed by the integrated optimization approach and implementation steps.

In this section, we introduce \OurModel{}, a systematic approach for generating adversarial objects that compromise the lift capability and grasp stability of robotic grasping. We first describe attack methods targeting these two grasp metrics individually, followed by the integrated framework and its implementation. 
Please refer to Fig.~\ref{fig:framework} for a demonstration.

%In this section, we present \OurModel{}, a comprehensive framework for systematically generating adversarial objects aimed at compromising the lift capability and grasp stability of robotic grasps. We first introduce attack methods targeting these two grasp metrics independently, followed by an integrated framework and its implementation details. An overview of the framework is illustrated in Fig.~\ref{fig:framework}.

\subsection{Adversarial Attack on Lift Capability}

\firstpara{Lift Capability of A Grasp}
The lift capability of a grasp reflects the robotic gripper's ability to counteract gravitational forces and achieve wrench equilibrium. As outlined in Section \ref{subsec:Stable}, this equilibrium requires the combined wrenches at all contact points to balance the gravitational wrench.

The general condition for wrench equilibrium described in Eqn.~\ref{eq:wrench_external} can be specified for gravity as:
\begin{equation}
\sum \mathbf{w}_i + \mathbf{w}_{gravity} =  \sum (\mathbf{f}_i, (\mathbf{c}_i - \mathbf{z}) \times \mathbf{f}_i + \tau_i) + (m\mathbf{g}, \mathbf{0}) = \mathbf{0},
\label{eq:wrench_gravity}
\end{equation}
where $m$ is the mass of the object, and $\mathbf{g}$ is the gravitational acceleration vector.

To quantify lift capability, we adopt the metric as formulated in \cite{ferrari1992planning}:
\begin{equation}
    LC(\mathcal{GSP}, \mathcal{O}) = \max_{\mathcal{F}_\perp \in \mathcal{G}(\mathbf{w}_{gravity})} \frac{\|\mathbf{w}_{gravity}\|}{\|\mathcal{F}_\perp\|},
\end{equation}
where the set \( \mathcal{G}(\mathbf{w}_{gravity}) \) contains the normal forces \( \mathcal{F}^\perp \) needed to balance the gravitational wrench \( \mathbf{w}_{gravity} \).

\if 0
\firstpara{Adversarial Strategy for Lift Capability}
We aim to reduce the lift capability \( LC \) by perturbing the object's geometry, e.g., its center of mass \( \mathbf{z} \). This optimization is formalized as:
\begin{equation}
\mathcal{O}' = \arg \min_{\mathcal{O}} LC(\mathcal{GSP}, \mathcal{O}),
\end{equation}
where \( \mathcal{O}' \) denotes the adversarially modified object. By altering the geometry of \( \mathcal{O} \), the resulting changes in \( \mathbf{z} \) increase the torques required for achieving wrench equilibrium, effectively lowering \( LC \).
\fi

\firstpara{Adversarial Strategy for Lift Capability}
We aim to reduce the lift capability \( LC \) by perturbing the object's geometry. This optimization is formalized as:
\begin{equation}
\mathcal{O}' = \arg \min_{\mathcal{O}} LC(\mathcal{GSP}, \mathcal{O}),
\end{equation}
where \( \mathcal{O}' \) represents the adversarially modified object. By altering the geometry of \( \mathcal{O} \), the gravitational torque is increased, necessitating a greater gripper force to maintain wrench equilibrium. This effectively reduces the object's lift capability \( LC \), making it more challenging for the gripper to lift the object.

\begin{figure*}[!t]
    \centering
\includegraphics[width=0.78\linewidth]{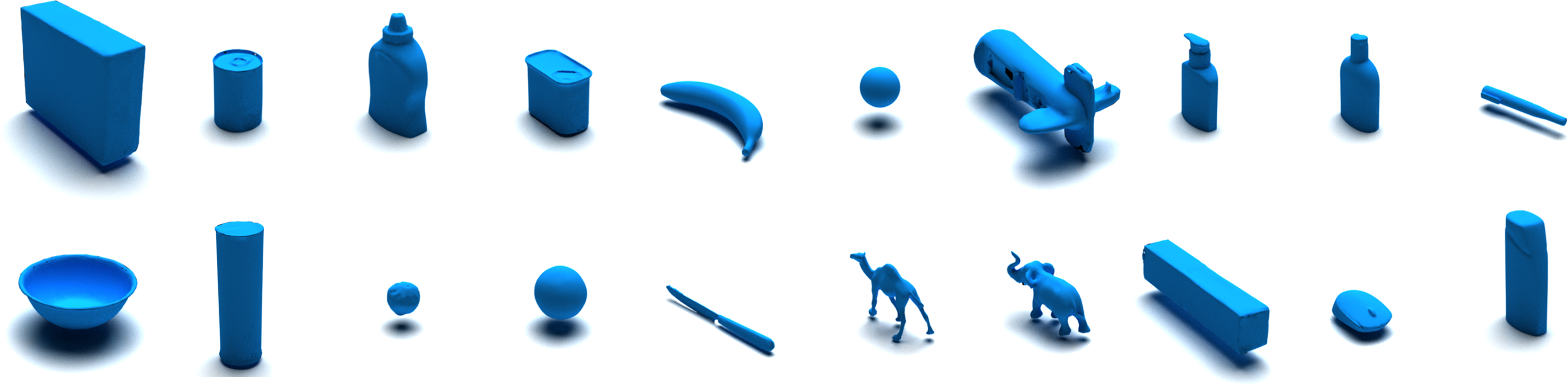}
    \caption{Visualization of the 20 objects in the AdvGrasp-20 benchmark. 
    The first row includes: \textsc{cracker box}, \textsc{tomato soup can}, \textsc{mustard bottle}, \textsc{potted meat can}, \textsc{banana}, \textsc{racquetball}, \textsc{toy airplane}, \textsc{wash soup}, \textsc{dabao sod}, \textsc{baoke marker}. The second row includes: \textsc{bowl}, \textsc{chips can}, \textsc{strawberry}, \textsc{orange}, \textsc{knife}, \textsc{camel}, \textsc{large elephant}, \textsc{darlie box}, \textsc{mouse}, \textsc{shampoo}. %Sample grasps for two-finger and three-finger grippers are also illustrated.
    }
    \label{fig:dataset}
\end{figure*}

\subsection{Adversarial Attack on Grasp Stability}

\firstpara{Grasp Stability of A Grasp}
Grasp stability measures the gripper's ability to resist external disturbances and maintain secure contact with the object throughout the manipulation process. 
As formulated in Ferrari \emph{et al.}~\shortcite{ferrari1992planning}, the stability score is expressed as:
\begin{equation}
    GS(\mathcal{GSP}, \mathcal{O}) = \min_{S \in \operatorname{ConvexHull}(\{\mathcal{W}_i\})} \operatorname{Dis_{p2s}}(\mathbf{0}, S),
\end{equation}
where \( \operatorname{ConvexHull}(\{\mathcal{W}_i\}) \) represents the convex hull of the wrenches \( \{\mathcal{W}_i\} \) applied at all contact points, and \( \operatorname{Dis_{p2s}} \) computes the distance between the origin and the convex hull.

\if 0
\firstpara{Adversarial Strategy for Grasp Stability}
We reduce the stability score \( GS \) by introducing deformations near the contact regions of the object, altering the normal vectors \( \mathbf{n}_i \) at these points. The optimization objective is:
\begin{equation}
\mathcal{O}' = \arg \min_{\mathcal{O}} GS(\mathcal{GSP}, \mathcal{O}).
\end{equation}
By intentionally introducing  deformations to the object's shape, which alter the surface normals \( \mathbf{n}_i \), the wrench space is adjusted, effectively reducing stability.
\fi

\firstpara{Adversarial Strategy for Grasp Stability}
We reduce the stability score \( GS \) by introducing deformations near the contact regions of the object, altering the normal vectors \( \mathbf{n}_i \) at these points. The optimization objective is:
\begin{equation}
\mathcal{O}' = \arg \min_{\mathcal{O}} GS(\mathcal{GSP}, \mathcal{O}).
\end{equation}
By introducing targeted deformations to the object's shape, the surface normals \( \mathbf{n}_i \) are altered, which adjusts the wrench space and reduces the stability margin. This effectively decreases the grasp's resistance to external disturbances, lowering the stability score \( GS \).

%increase the gravitational torque and reduce the stability margin in the wrench space,

%By introducing strategic deformations to the object's surface normals \( \mathbf{n}_i \), the wrench space is adjusted, effectively reducing stability.

\if 0
\subsection{Unified \OurModel{} Framework}
Our method provides a unified framework that simultaneously addresses both lift capability and grasp stability. The optimization objective is defined as:
\begin{equation}
\mathcal{O}' = \arg \min_{\mathcal{O}} LC(\mathcal{GSP}, \mathcal{O}) + \lambda_1 GS(\mathcal{GSP}, \mathcal{O}) + \lambda_2 \operatorname{Lap}(\mathcal{O}),
\label{eq:unified_objective}
\end{equation}
where \( \operatorname{Lap}(\mathcal{O}) \) ensures smooth geometric deformations by penalizing irregularities in the object shape:
\begin{equation}
\operatorname{Lap}(\mathcal{O}) = \sum_{v \in \mathcal{V}(\mathcal{O})} \left( \frac{1}{|\operatorname{Neigh}(v)|} \sum_{v' \in \operatorname{Neigh}(v)} (v' - v) \right)^2.
\end{equation}
Here, \( \mathcal{V}(\mathcal{O}) \) represents the vertices of \( \mathcal{O} \), and \( \operatorname{Neigh}(v) \) denotes the neighboring vertices of \( v \).

This unified approach effectively integrates the two critical aspects of robotic grasping performance, ensuring that the adversarial modifications are both comprehensive and physically plausible.
\fi

\subsection{Unified \OurModel{} Framework}
Our method provides a unified framework that simultaneously addresses lift capability and grasp stability, ensuring comprehensive adversarial modifications. The optimization objective is defined as:
\begin{equation}
\mathcal{O}' = \arg \min_{\mathcal{O}} LC(\mathcal{GSP}, \mathcal{O}) + \lambda_1 GS(\mathcal{GSP}, \mathcal{O}) + \lambda_2 \operatorname{Lap}(\mathcal{O}),
\label{eq:unified_objective}
\end{equation}
where \( \operatorname{Lap}(\mathcal{O}) \) ensures smooth geometric deformations by penalizing irregularities in the object shape:
\begin{equation}
\operatorname{Lap}(\mathcal{O}) = \sum_{v \in \mathcal{V}(\mathcal{O})} \left( \frac{1}{|\operatorname{Neigh}(v)|} \sum_{v' \in \operatorname{Neigh}(v)} (v' - v) \right)^2.
\end{equation}
%Here, \( \mathcal{V}(\mathcal{O}) \) represents the vertices of \( \mathcal{O} \), \( \operatorname{Neigh}(v) \) denotes the neighboring vertices of \( v \), and \( \lambda_1 \) and \( \lambda_2 \) are weighting parameters.
Here, \( \mathcal{V}(\mathcal{O}) \) represents the vertices of \( \mathcal{O} \), \( \operatorname{Neigh}(v) \) denotes the neighboring vertices of \( v \), and \( \lambda_1 \) and \( \lambda_2 \) are hyperparameters that balance the contributions of grasp stability, lift capability, and deformation regularization, respectively.

%Here, \( \mathcal{V}(\mathcal{O}) \) represents the vertices of \( \mathcal{O} \), and \( \operatorname{Neigh}(v) \) denotes the neighboring vertices of \( v \), and \lambda_1 and \lambda_2  are weighting parameters.

\subsection{Implementation of \OurModel{}}

\OurModel{} generates adversarial objects \( \mathcal{O}' \) for robotic grasping through the following three steps:

\firstpara{Bounding Box Initialization}
An axis-aligned bounding box  is constructed for the object \( \mathcal{O} \), providing spatial constraints for defining deformation regions.

\firstpara{Control Point Placement}
Control points are evenly  distributed across the bounding box surfaces, creating a structured framework for guiding subsequent deformations.

\firstpara{Iterative Shape Deformation and Optimization}
Using the control points as anchors, iterative perturbations are applied to the object's geometry following~\cite{ju2023mean}. In each iteration, the resolution of control points is progressively increased, allowing for finer adjustments. 

The adversarial object \( \mathcal{O}' \) is refined to maximize degradation of grasp performance, while ensuring physical plausibility and maintaining imperceptibility of the modifications.

\begin{table*}[!t]
\centering
\setlength{\tabcolsep}{0.7mm}{
\scalebox{0.84}{
\begin{tabular}{c|cccccccc|cccccccc}
\hline
\multirow{3}{*}{Object}               & \multicolumn{8}{c|}{2-Finger Grasp}                                                                                                                      & \multicolumn{8}{c}{3-Finger Grasp}                                                                                                                       \\ \cline{2-17} 
                                      & \multicolumn{4}{c|}{MinGF}                                                             & \multicolumn{4}{c|}{MaxLM}                                      & \multicolumn{4}{c|}{MinGF}                                                             & \multicolumn{4}{c}{MaxLM}                                       \\ \cline{2-17} 
                                      & Origin & ALC           & AGS           & \multicolumn{1}{c|}{\OurModel{}} & Origin & ALC          & AGS          & \OurModel{} & Origin & ALC           & AGS           & \multicolumn{1}{c|}{\OurModel{}} & Origin & ALC          & AGS          & \OurModel{} \\ \hline
\textsc{cracker box}     & 32.3   & \textbf{44.4} & 34.8          & \multicolumn{1}{c|}{43.6}                     & 2.8    & \textbf{2.6} & 2.6          & 2.7                      & 16.8   & 11.2          & 8.4           & \multicolumn{1}{c|}{12.2}                     & 2.8    & 1.7          & 2.7          & \textbf{1.5}             \\
\textsc{tomato soup can} & 16.2   & 17.1          & \textbf{33.0} & \multicolumn{1}{c|}{32.0}                     & 3.7    & \textbf{2.6} & 3.2          & 3.3                      & 8.8    & 17.0          & 11.6          & \multicolumn{1}{c|}{\textbf{29.4}}            & 9.9    & 4.0          & 4.1          & \textbf{3.5}             \\
\textsc{mustard bottle}  & 22.6   & \textbf{26.0} & 12.8          & \multicolumn{1}{c|}{20.7}                     & 3.9    & 3.9          & \textbf{3.7} & 3.8                      & 12.0   & 13.2 & 10.0          & \multicolumn{1}{c|}{\textbf{14.5}}                     & 5.7    & \textbf{3.6} & 4.6          & 3.8                      \\
\textsc{potted meat can} & 10.6   & \textbf{20.6} & 11.6          & \multicolumn{1}{c|}{11.9}                     & 3.9    & 3.8          & 3.8          & \textbf{3.8}             & 10.6   & \textbf{19.4} & 9.6           & \multicolumn{1}{c|}{10.0}                     & 4.6    & 3.8          & \textbf{2.9} & 3.2                      \\
\textsc{banana}          & 16.6   & \textbf{36.7} & 27.4          & \multicolumn{1}{c|}{29.9}                     & 3.7    & \textbf{1.9} & 4.0          & 3.3                      & 11.0   & 12.6          & 17.6          & \multicolumn{1}{c|}{\textbf{20.1}}            & 9.6    & 6.0          & \textbf{2.1} & 4.9                      \\
\textsc{bowl}            & 32.5   & \textbf{33.4} & 27.2          & \multicolumn{1}{c|}{31.3}                     & 3.4    & 3.1          & \textbf{3.1} & 3.2                      & 5.8    & \textbf{16.6} & 6.0           & \multicolumn{1}{c|}{7.8}                      & 11.8   & 11.2         & 7.1          & \textbf{6.4}             \\
\textsc{chips can}       & 27.9   & 34.0          & \textbf{34.7} & \multicolumn{1}{c|}{28.5}                     & 3.3    & \textbf{2.4} & 2.5          & 3.0                      & 16.8   & 10.7          & 8.8           & \multicolumn{1}{c|}{\textbf{11.3}}                     & 5.8    & 3.5          & \textbf{2.4} & 3.5                      \\
\textsc{strawberry}      & 8.3    & 31.7          & 22.8          & \multicolumn{1}{c|}{\textbf{33.2}}            & 4.3    & \textbf{4.1} & 4.3          & 4.1                      & 10.8   & 12.2          & \textbf{15.4} & \multicolumn{1}{c|}{11.8}                     & 6.4    & \textbf{1.5} & 2.0          & 2.1                      \\
\textsc{orange}          & 25.4   & \textbf{33.6} & 26.3          & \multicolumn{1}{c|}{27.8}                     & 2.4    & \textbf{1.6} & 2.4          & 1.8                      & 10.8   & \textbf{12.6} & 10.8          & \multicolumn{1}{c|}{10.8}                     & 4.9    & \textbf{1.8} & 1.9          & 2.1                      \\
\textsc{knife}           & 9.9    & 30.4          & 37.9          & \multicolumn{1}{c|}{\textbf{39.9}}            & 4.0    & 4.2          & 3.7          & \textbf{3.5}             & 11.0   & 11.6          & 11.2          & \multicolumn{1}{c|}{\textbf{12.8}}            & 4.0    & 4.0          & 3.7          & \textbf{3.7}             \\
\textsc{racquetball}     & 21.9   & 7.4           & 14.6          & \multicolumn{1}{c|}{\textbf{20.8}}            & 3.9    & 5.0          & 4.1          & \textbf{2.1}             & 10.6   & 21.2          & 24.3          & \multicolumn{1}{c|}{\textbf{36.0}}            & 4.9    & 2.0          & 2.4          & \textbf{1.6}             \\
\textsc{toy airplane}    & 29.2   & 11.9          & \textbf{22.1} & \multicolumn{1}{c|}{14.6}                     & 3.6    & 4.6          & 3.5          & \textbf{1.8}             & 22.0   & 49.6          & 17.9          & \multicolumn{1}{c|}{\textbf{50.0}}            & 3.7    & 3.7          & \textbf{2.0} & 2.0                      \\
\textsc{wash soup}       & 8.5    & \textbf{32.8} & 24.9          & \multicolumn{1}{c|}{26.0}                     & 3.0    & 2.9          & \textbf{2.2} & 3.1                      & 12.2   & \textbf{19.6} & 13.8          & \multicolumn{1}{c|}{14.2}                     & 3.3    & 1.4          & \textbf{1.4} & 1.8                      \\
\textsc{dabao sod}       & 27.9   & \textbf{34.1} & 32.5          & \multicolumn{1}{c|}{32.8}                     & 3.7    & \textbf{3.0} & 3.2          & 3.7                      & 10.8   & 26.2          & \textbf{31.9} & \multicolumn{1}{c|}{14.0}                     & 3.4    & \textbf{1.9} & 2.6          & 2.6                      \\
\textsc{baoke marker}    & 17.4   & 26.4          & 23.8          & \multicolumn{1}{c|}{\textbf{33.9}}            & 3.9    & \textbf{3.3} & 3.9          & 3.9                      & 18.2   & 18.0          & 16.2          & \multicolumn{1}{c|}{\textbf{19.8}}            & 2.8    & \textbf{1.9} & 2.0          & 2.1                      \\
\textsc{camel}           & 15.5   & 19.8          & \textbf{22.7} & \multicolumn{1}{c|}{9.7}                      & 4.0    & \textbf{3.7} & 3.9          & 3.9                      & 6.4    & \textbf{12.0}          & 5.3           & \multicolumn{1}{c|}{10.4}            & 7.2    & 6.5          & \textbf{3.6} & 4.3                      \\
\textsc{large elephant}  & 16.8   & \textbf{27.0} & 23.8          & \multicolumn{1}{c|}{19.5}                     & 3.9    & 4.0          & \textbf{3.8} & 3.9                      & 12.2   & \textbf{26.4} & 19.6          & \multicolumn{1}{c|}{23.4}                     & 5.3    & 2.6 & 2.9          & \textbf{2.3}                      \\
\textsc{darlie box}      & 14.8   & 26.6          & 26.7          & \multicolumn{1}{c|}{\textbf{26.8}}            & 3.8    & \textbf{2.6} & 3.1          & 3.0                      & 6.0    & 19.8          & 16.2          & \multicolumn{1}{c|}{\textbf{24.6}}            & 5.3    & \textbf{1.4} & 1.6          & 1.8                      \\
\textsc{mouse}           & 20.7   & \textbf{27.3} & 22.8          & \multicolumn{1}{c|}{25.4}                     & 3.6    & \textbf{1.5} & 2.4          & 1.9                      & 13.2   & 26.6          & 25.2          & \multicolumn{1}{c|}{\textbf{27.0}}            & 4.5    & \textbf{4.2} & 4.2          & 4.4                      \\
\textsc{shampoo}         & 21.9   & 24.9          & \textbf{29.6} & \multicolumn{1}{c|}{14.0}                     & 3.3    & 2.3          & \textbf{1.9} & 2.2                      & 10.8   & 27.1          & 25.6          & \multicolumn{1}{c|}{\textbf{38.8}}            & 9.6    & \textbf{3.1} & 6.8          & 4.5                      \\ \hline
\end{tabular}
}}
\caption{Comparison of attack performance in counteracting gravity for two-finger and three-finger robotic grasping.}
\label{tab:grasp_gravity}
\end{table*}

\section{AdvGrasp-20 Benchmark}

To advance research in the emerging field of adversarial attacks on robotic grasping, we establish a comprehensive benchmark, AdvGrasp-20, designed to facilitate fair comparisons and promote further development in this area.

\firstpara{Object Selection}
We select 20 representative objects with diverse shapes from the GraspNet-1Billion dataset~\cite{fang2020graspnet,fang2023graspnet2}, including common items such as \textsc{cracker box}, \textsc{tomato soup can}, and \textsc{mustard bottle}. 
Please refer to Fig.~\ref{fig:dataset} for a visual demonstration.

\firstpara{Grasp Generation}
For each object, we generate 5 grasps for a two-finger gripper using the analytic Dex-Net 2.0 framework~\cite{Mahler-DexNet} and 5 grasps using the deep learning-based GraspNet. In addition, we generate 5 grasps for a three finger gripper using deep learning-based GenDexGrasp~\cite{li2022gendexgrasp}. 

\firstpara{Post-processing}
To ensure the generated grasps are feasible, we further perform post-processing. Specifically, we validate the grasps in the PyBullet simulator using the Robotiq 2-finger 2F-85 gripper and the Robotiq 3-finger gripper. Only the grasps that successfully lift the objects are retained as final grasps, while unsuccessful ones are filtered out.

By including diverse objects and grasp strategies, AdvGrasp-20 provides a robust platform for systematically evaluating adversarial attacks on various grasping configurations. This benchmark not only enables fair comparisons but also fosters further innovation in the field of robotic grasping under adversarial conditions.

\section{Experimental Results}

\subsection{Experimental Setup}

\firstpara{Implementation}
We implement our framework in Python using the PyBullet simulator~\cite{coumans2016pybullet} as the simulation environment.  
All simulations employ the soft finger contact model~\cite{mahler2018dex}, with the friction coefficient set to \( \mu = 0.6 \) and the torsional friction coefficient \( \gamma = 0.3 \).  
To manipulate the object's shape, each face's four edges are evenly divided into a square grid based on the cage size, initially set to 0.04. The vertices of these grids serve as control points, which act as anchors for shape manipulation. The optimization process utilizes a simulated annealing algorithm, with an initial temperature \( T_0 = 1000 \), a minimum temperature \( T_{\text{min}} = 10^{-5} \), a cooling rate \( \alpha = 0.98 \), and a perturbation scale \( \epsilon = 0.05 \times \text{cage size} \) for control points.  
After completing each optimization cycle, the cage size is halved, and new control points are generated for further optimization. This process is repeated for a total of five cycles.  
To balance the contributions of lift capability, grasp stability, and shape regularization in AdvGrasp, we set the weighting parameters to \( \lambda_1 = 10000 \) and \( \lambda_2 = 50 \).

\firstpara{Our Attack Solutions}
We consider three attack solutions for evaluation: adversarial attacks targeting only lift capability (\textbf{ALC}), adversarial attacks targeting only grasp stability (\textbf{AGS}), and the unified approach, \textbf{\OurModel{}}, which simultaneously considers both aspects.

%\firstpara{Evaluation Metrics}
%Unlike adversarial attacks in image classification, where success is determined by a straightforward binary outcome, evaluating adversarial attacks on robotic grasping from a physical perspective is significantly more complex. 

%\firstpara{Evaluation Metrics} Unlike the binary success criteria in image classification adversarial attacks, evaluating adversarial attacks on robotic grasping from a physical perspective is far more complex. Therefore, {\textbf{we define grasp failure as a scenario where the object experiences significant slippage}, which is explicitly defined as the displacement exceeding a threshold of 0.02 m or the rotation exceeding \( 10^\circ \). Based on this definition, we devise three metrics to evaluate the performance of attack robotic grasping:

\firstpara{Evaluation Metrics}
Unlike the binary success criteria in image classification adversarial attacks, evaluating adversarial attacks on robotic grasping from a physical perspective is inherently more complex. To address this, \textbf{we define a grasp failure as a scenario where the object undergoes significant slippage}, explicitly characterized by a displacement exceeding 0.02~m or a rotation surpassing \(10^\circ\). Based on this definition, we establish three metrics to evaluate the performance of adversarial attacks on robotic grasping:
\begin{itemize}
\setlength{\itemsep}{0pt}
\setlength{\parsep}{0pt}
\setlength{\parskip}{0pt}
\item  \textbf{Minimal Grasp Force (MinGF)}:  
The minimal force required by the gripper to lift a 1 kg object. Specifically, we start the grasp force at 50 N and decrease it by 0.2 N per step until significant slippage occurs.

\item  \textbf{Maximal Lifting Mass (MaxLM)}:  
The maximum weight the gripper can lift when each finger applies a maximum force of 50 N. Specifically, we start the object's weight at 1 kg and increase it by 0.1 kg per step until significant slippage occurs.

\item  \textbf{Maximal External Disturbance (MaxED)}:  
The maximum external force a grasp can withstand while holding a 1 kg object, with each finger of the gripper applying a maximum force of 50 N. We use Fibonacci sphere sampling~\cite{larkins2012analysis} to generate 50 directions on a unit sphere. For each direction, a force is applied over a time interval of \( 0.008 \, \mathrm{s} \), starting at 1 N and increasing by 1 N per step. The force is incremented only if no significant slippage occurs in all 50 directions.
\end{itemize}

Among these metrics, MinGF and MaxLM primarily evaluate the grasp configuration's ability to counteract gravity, while MaxED focuses on its resistance to external forces. Together, these metrics form a comprehensive framework for assessing the physical robustness of robotic grasp configurations under adversarial conditions.

\subsection{Main Results and Analyses}

\begin{figure*}[!t]
    \centering
\includegraphics[width=0.93\linewidth]{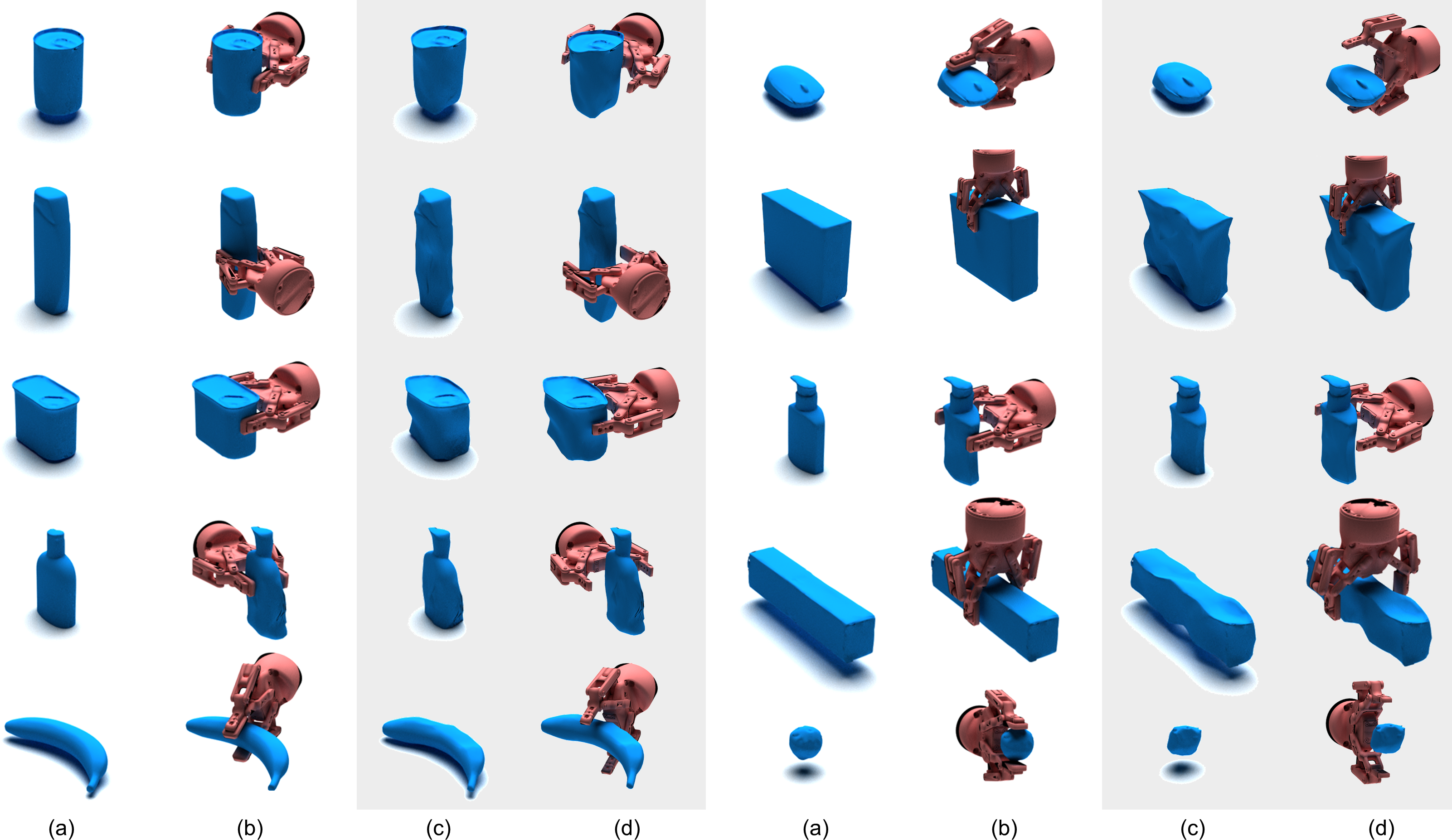}
    \caption{Visualization of the effects of \OurModel{}. (a) The original object, (b) the grasp configuration on the original object, (c) the adversarial object generated by \OurModel{}, and (d) the failed grasp on the adversarial object due to modifications introduced by \OurModel{}.
    }
    \label{fig:simulation}
\end{figure*}

\begin{table}[!t]
\centering
\setlength{\tabcolsep}{0.6mm}{
\scalebox{0.72}{
\begin{tabular}{c|cccc|cccc}
    \hline
    \multirow{2}{*}{Object}               & \multicolumn{4}{c|}{2-Finger Grasp}                           & \multicolumn{4}{c}{3-Finger Grasp}                            \\ \cline{2-9} 
                                          & Origin & ALC         & AGS         & \OurModel{} & Origin & ALC         & AGS         & \OurModel{} \\ \hline
    \textsc{cracker box}     & 29     & \textbf{18} & 23          & 28                       & 24     & \textbf{29} & 32          & 36                       \\
    \textsc{tomato soup can} & 33     & 14          & \textbf{14} & 24                       & 49     & 31          & 28          & \textbf{25}              \\
    \textsc{mustard bottle}  & 26     & \textbf{23} & 23          & 27                       & 37     & \textbf{21} & 24          & 26                       \\
    \textsc{potted meat can} & 37     & 31          & 32          & \textbf{28}              & 51     & 38          & 35          & \textbf{20}              \\
    \textsc{banana}          & 56     & 33          & 32          & \textbf{19}              & 62     & 42          & 48          & \textbf{32}              \\
    \textsc{bowl}            & 36     & 33          & \textbf{27} & 32                       & 68     & 57          & 60          & \textbf{24}              \\
    \textsc{chips can}       & 33     & \textbf{21} & 34          & 35                       & 63     & \textbf{34} & 41          & 37                       \\
    \textsc{strawberry}      & 41     & \textbf{28} & 30          & 31                       & 72     & 44          & \textbf{35} & 36                       \\
    \textsc{orange}          & 14     & \textbf{7}  & 9           & 8                        & 67     & 36          & 38          & \textbf{33}              \\
    \textsc{knife}           & 33     & 20          & 29          & \textbf{10}              & 7      & 17          & 23          & \textbf{13}              \\
    \textsc{racquetball}     & 21     & 48          & 21          & \textbf{17}              & 53     & \textbf{24} & 35          & 27                       \\
    \textsc{toy airplane}     & 27     & 35          & 36          & \textbf{16}             & 44     & 32          & 44          & \textbf{21}              \\
    \textsc{wash soup}       & 32     & \textbf{15} & 20          & 16                       & 54     & 45          & 32          & \textbf{24}              \\
    \textsc{dabao sod}       & 34     & 10          & \textbf{5}  & 20                       & 31     & 33          & \textbf{21} & 24                       \\
    \textsc{baoke marker}    & 37     & 26          & 29          & \textbf{21}              & 15     & \textbf{27} & 34          & 40                       \\
    \textsc{camel}           & 37     & 32          & \textbf{25} & 40                       & 23     & 21          & \textbf{19} & 27                       \\
    \textsc{large elephant}  & 36     & 30          & \textbf{19} & 23                       & 33     & \textbf{18} & 26          & 30                       \\
    \textsc{darlie box}      & 50     & 31          & 19          & \textbf{17}              & 31     & \textbf{23} & 33          & 25                       \\
    \textsc{mouse}           & 25     & 21          & 6           & \textbf{1}               & 35     & 29          & \textbf{28} & 28                       \\
    \textsc{shampoo}         & 19     & 22          & \textbf{7}  & 19                       & 41     & \textbf{15} & 17          & 20                       \\ \hline
    \end{tabular}
}
}
\caption{
%Comparison of attack performance in resisting external forces for two-finger and three-finger robotic grasping measured by MaxED.
Comparison of attack performance in resisting external forces for two- and three-finger  grasping, measured by MaxED.
}
\label{tab:external_forces}
\end{table}

\firstpara{Attack Performance on Counteracting Gravity}
As shown in Tab.~\ref{tab:grasp_gravity}, for objects of the same type and mass (1 kg), three-finger grasps generally achieve a lower MinGF compared to two-finger grasps, indicating that three-finger grippers require less force to counteract gravity for the same object. Additionally, under the same condition where each finger can apply a maximum force of 50 N, three-finger grippers can often counteract gravity for objects with greater mass.

After applying our method, the effectiveness of the original grasp configurations significantly decreases, as reflected by increased MinGF and decreased MaxLM. For most objects, ALC proves more effective in making counteracting gravity more difficult, as it introduces gravity-related factors into the optimization. In contrast, AGS is less effective in this regard.

%\wang{update knife racquetball toyairplane data}

\begin{figure}[!t]
    \centering
    \includegraphics[width=0.85\linewidth]{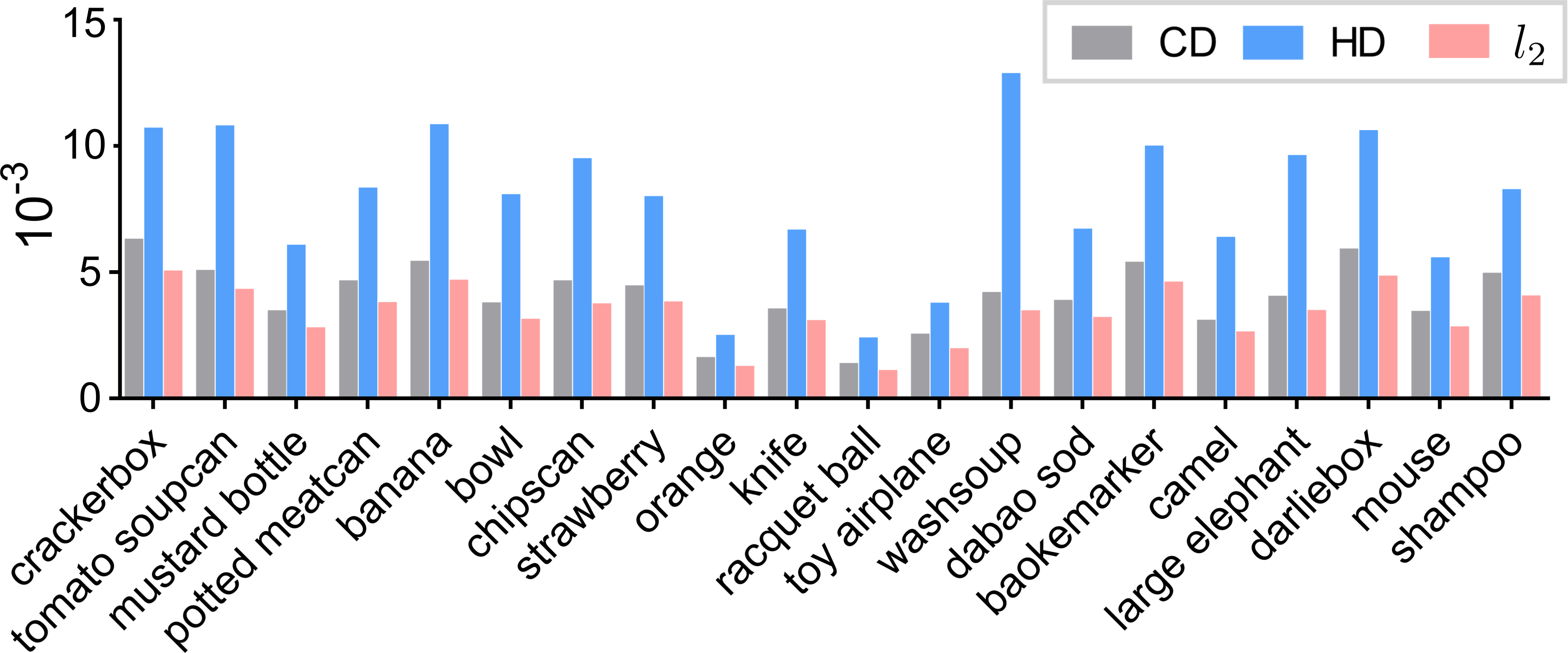}
    \caption{The deformation introduced by \OurModel{}, measured by 
    Chamfer distance (CD), Hausdorff distance (HD), and \(l_2\)-norm (\(l_2\))   computed on the mesh vertices between the adversarial objects and the original objects.
    }
    \label{fig:distance}
\end{figure}

\firstpara{Attack Performance on Resisting External Forces}
We further evaluate the ability of these grasps to resist external forces under attack.
As shown in Tab.~\ref{tab:external_forces}, three-finger grasps generally tolerate higher disturbances compared to two-finger grasps, indicating greater robustness to external perturbations. 
However, after applying our attacks, particularly the unified \OurModel{}, the stability of these grasps is significantly compromised, making them more vulnerable to disruption. These results validate the effectiveness of our  methods.

%However, after applying our attacks, 特别是unified \OurModel{} the stability of these grasps is significantly reduced, making them more susceptible to disruption. These results validate the effectiveness of our proposed methods.

% 目前的单位是牛顿，要不要更新成冲量单位？部分数据需要更新
% 从表格中的横行可以看出三爪的抓取往往比二爪能够容忍更高的扰动，也就是具有更高的抗扰动稳定性。相较于原始的mesh，通过我们的三种方法得到的对抗mesh更容易被扰动破坏抓取的稳定性，说明了我们提出的算法的有效性。

\begin{figure*}[!t]
    \centering
    \includegraphics[width=0.93\linewidth]{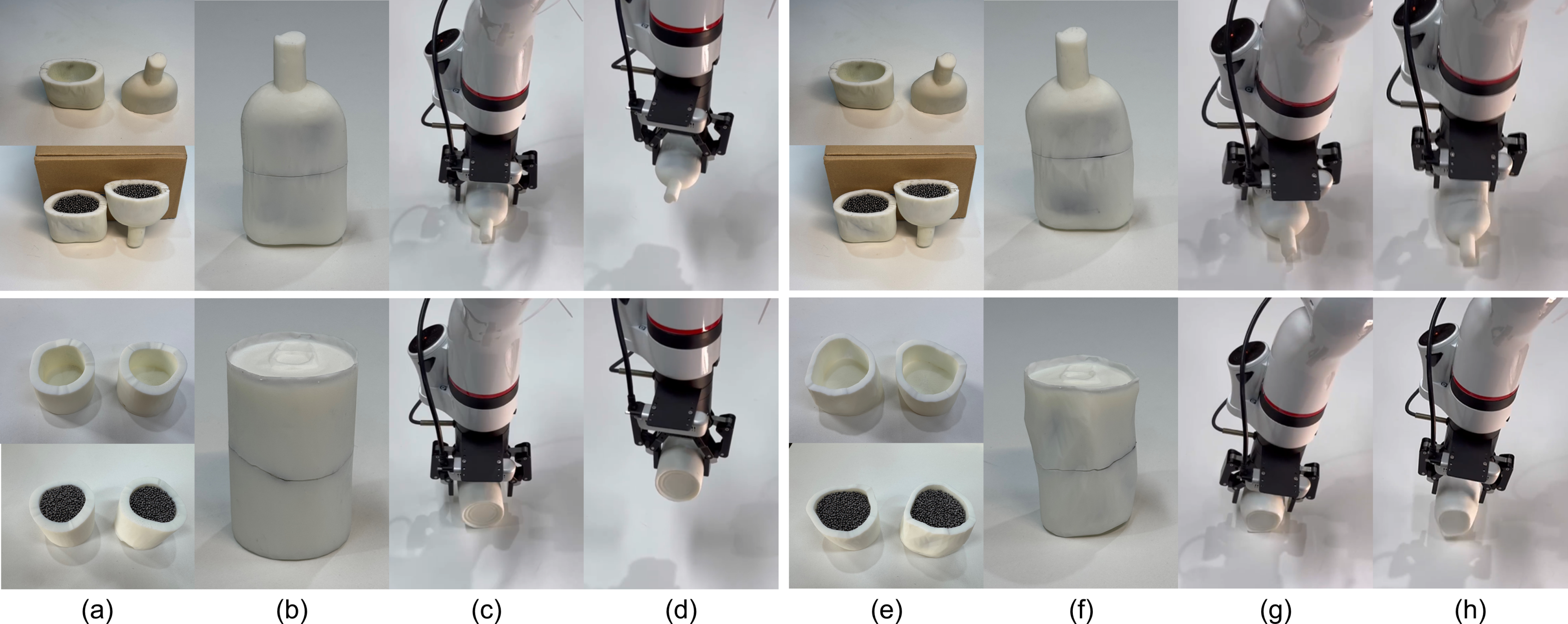}
    \caption{
Visualization of the adversarial attack  on robotic grasping in a physical scenario: (a) assembly process with lead balls, (b) assembled 3D-printed object, (c) gripper preparing to grasp, and (d) grasping result. The corresponding adversarial process generated by \OurModel{}: (e) adversarial assembly process, (f) assembled adversarial object, (g) gripper preparing to grasp the adversarial object, and (h) adversarial grasping result, which fails due to instability. The detailed process is available in the video provided in the supplementary materials.
    %Visualization of the physical process for the original grasp: (a) assembly process with lead balls, (b) assembled 3D-printed object, (c) gripper preparing to grasp, and (d) grasping result. The corresponding adversarial process generated by \OurModel{}: (e) adversarial assembly process, (f) assembled adversarial object, (g) gripper preparing to grasp the adversarial object, and (h) adversarial grasping result.
    }
    \label{fig:physics}
\end{figure*}

\begin{figure}[!t]
    \centering
    \includegraphics[width=0.63\linewidth]{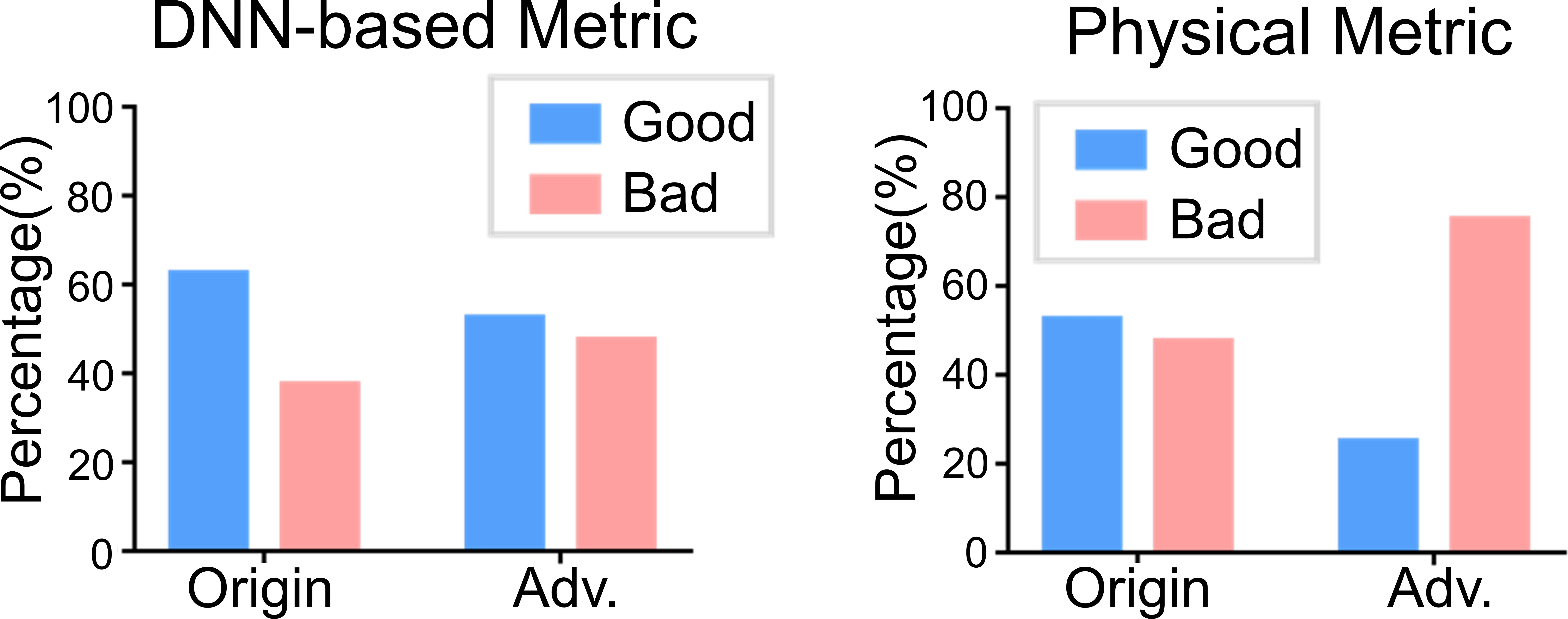}
    \caption{Proportion of original and adversarial grasps generated by \OurModel{} classified as good or bad, evaluated using both the DNN-based metric and the physical metric.
}
    \label{fig:pointgpd stats}
\end{figure}

\firstpara{Visualization}
Fig.~\ref{fig:simulation} provides a visual comparison between the original objects with their corresponding grasps and the adversarial objects with their respective counterparts. It shows that the overall shape of the objects is preserved. At the same time, localized deformations occur near the grasp contact points, including changes in surface normals and slight shifts in the center of mass. These perturbations disrupt the grasp's wrench equilibrium, validating the effectiveness of our method. This demonstrates how \OurModel{} generates imperceptible yet impactful adversarial perturbations to effectively compromise robotic grasping performance.

%Fig.~\ref{fig:simulation} provides a visual comparison between the original objects with their corresponding grasps and the adversarial objects with their respective counterparts. 此外，, the visualizations highlight localized deformations near the grasp points. These include changes in surface normals and slight shifts in the center of mass. Such modifications disrupt the grasp's physical equilibrium, validating the working mechanism of our method. This demonstrates how \OurModel{} effectively generates imperceptible yet impactful adversarial perturbations to compromise robotic grasping performance.

\if 0
\firstpara{Analysis on Shape Deformation}
We analyze the impact of \OurModel{} on the object's geometry quantitatively. As illustrated in Fig.~\ref{fig:distance},  confirm that the perturbations remain minimal, 一定程度上aligning with the requirements of imperceptibility in adversarial attacks. These results highlight the effectiveness of \OurModel{} in achieving adversarial objectives while maintaining geometric plausibility.
\fi

\firstpara{Analysis on Shape Deformation}
We quantitatively analyze the impact of \OurModel{} on the object's geometry. As illustrated in Fig.~\ref{fig:distance}, the perturbations remain minimal, aligning to some extent with the imperceptibility requirements of adversarial attacks. These results underscore the effectiveness of \OurModel{} in achieving adversarial objectives while preserving geometric plausibility.

%\wang{TODO: Update the style of the bar chart and add desprittion and refer to the figure in Chinese}
% 对于实际体积较大的物体，相较于较小的物体，生成的对抗样本的形变也会更大一些。

\subsection{Additional  Analyses}

\firstpara{Physical Results}
To validate the effectiveness of our method in real-world scenarios, we select the {\textsc{dabao sod}} and {\textsc{tomato soup can}} for physical experiments. For each object, we choose a specific grasp configuration and generate adversarial versions  using AdvGrasp.
We 3D-print these objects and modify them by hollowing them out, splitting them into two halves, and inserting lead balls to increase their mass to 1 kg, addressing the lightweight nature of 3D-printed materials.
Grasping experiments are conducted under controlled conditions using a Robotiq 2-finger 2F-85 gripper mounted on a robotic arm. When a maximum grasping force of 30 N is applied to the \textsc{dabao sod} and its adversarial version, the original model is successfully lifted, while the adversarial model  exhibits slippage and subsequently drops. Similarly, when a 20 N grasping force is applied to the \textsc{tomato soup can} and its adversarial version, the original model is successfully lifted, while the adversarial model  fails and is dropped.
These experiments highlight the impact of physical properties on grasp quality and confirm the effectiveness of AdvGrasp in generating adversarial objects that compromise grasp stability in real-world conditions.

\firstpara{DNN-based Metric vs. Physical Metric}
To validate the importance of physical metrics in grasp evaluation, we assess all grasps in the AdvGrasp-20 benchmark, as well as adversarial grasps generated by \OurModel{}, using both the DNN-based PointNetGPD~\cite{liang2019pointnetgpd} and  physical metric. Specifically, we apply the 2-class classification approach from PointNetGPD to determine whether a grasp is good or bad.
For the physical metric, a grasp is considered bad if the actual grasp position deviates from the expected position by more than 0.03 m or if the angle deviation exceeds 10 degrees during simulation experiments.
As shown in Fig.~\ref{fig:pointgpd stats}, the physics-based metric identifies most adversarial grasps 
generated by \OurModel{} as bad. However, PointNetGPD, which relies solely on local geometric properties, fails to accurately assess these cases.
These findings underscore the indispensable role of physical metrics in grasp evaluation.

\if 0
\firstpara{DNN-based Metric vs. Physical Metric}
To validate the importance of physical properties in grasp evaluation, we assess all grasps in the AdvGrasp-20 Benchmark, as well as adversarial grasps generated by \OurModel{}, using both the DNN-based PointNetGPD~\cite{liang2019pointnetgpd} and our proposed physics-based metric. Specifically, we apply the 2-class classification approach from PointNetGPD to dertemine grasp good or bad.
For the physical metric, a grasp is considered unsuccessful if the actual grasp position deviates from the expected position by more than 3 cm or if the angle deviation exceeds 10 degrees during simulation experiments.
As shown in Fig.~\ref{fig:pointgpd stats}, 
经过了\OurModel{}的强力对抗攻击，物理based的metric显示确实大部分对抗下的抓取是一个bad，但是
，PointNetGPD which relies solely on local point cloud data, fails to accurately assess it。
hese findings highlight the indispensability role of physical metrics in grasp evaluation.
\fi
%demonstrating their indispensability for accurate and robust assessments.

%simulation results reveal that many grasps suffer quality degradation due to changes in physical attributes, such as object shape. In contrast, PointNetGPD, which relies solely on local point cloud data, fails to accurately assess grasp quality due to its lack of physical context. These findings highlight the critical role of physical metrics in grasp evaluation, demonstrating their indispensability for accurate and robust assessments.

\section{Conclusion}

In this paper, we have proposed a framework for adversarial attacks on robotic grasping, targeting lift capability and grasp stability. Through shape deformation to increase gravitational torque and reduce stability margin in the wrench space, our method systematically degrades grasp performance. Extensive experiments in simulated and real-world environments validate the effectiveness of our approach. 
Future work will consider dynamic settings and focus on defense design.

%We hope our work can serve as a benchmark to inspire more research in this area. 
%Future work will explore dynamic scenarios and focus on developing defense mechanisms to enhance robustness.

\section*{Acknowledgements}
This work was supported in part by 
the Dreams Foundation of Jianghuai Advance Technology Center (2023-ZM01G003),
the National Natural Science Foundation of China (62472117), the Guangdong Basic and Applied Basic Research Foundation (2025A1515010157), and the Science and Technology Projects in Guangzhou (2025A03J0137).
% Dreams Foundation of Jianghuai Advance Technology Center (No. 2023-ZM01G003)
%In this paper, we have proposed a framework for adversarial attacks on robotic grasping, focusing on lift capability and grasp stability. By perturbing the object's center of mass and surface normals, our method systematically degrades grasp performance. Experimental results demonstrate the effectiveness of our approach in both simulated and physical environments. Future work will investigate extensions to dynamic scenarios and the development of defense mechanisms.

\bibliographystyle{named}
\bibliography{ijcai25}

% \bibliography{ijcai25_templateref}

\end{document}